\documentclass[letterpaper, 10 pt, conference]{ieeeconf}  

\usepackage{epsfig}
\usepackage{graphicx}
\usepackage{amsmath}
\usepackage{amssymb}
\usepackage[ruled, linesnumbered]{algorithm2e}
\usepackage{booktabs}
\usepackage{multirow}
\usepackage{color}
\usepackage{bm}
\usepackage[mathscr]{euscript}
\usepackage{float}
\usepackage{verbatim}
\usepackage{authblk}
\usepackage{url}
\usepackage[most]{tcolorbox}
\usepackage{subcaption}

\newtcolorbox[auto counter, number within=section]{mybox}[1][Example Box]{
	left=0pt,
	colback=white,
	colframe=blue!75!black,
	title=#1,
	width=0.5\textwidth,
	breakable,
	enhanced,
	arc=1mm,
	auto outer arc,
	title=#1,
	label={box:#1} 
}

\IEEEoverridecommandlockouts                              

\title{\LARGE \bf
	General-Purpose Aerial Intelligent Agents Empowered by \\ Large Language Models
}

\author{Ji Zhao and Xiao Lin
	\thanks{X. Lin is with the Institute of Automation, Chinese Academy of Sciences, Beijing 100190, China. X. Lin is the corresponding author.
	Email: {\tt\small xiao.lin@@ia.ac.cn}}
}

\begin{document}
	
	\maketitle
	\thispagestyle{empty}
	\pagestyle{empty}
	
	\begin{abstract}
		The emergence of large language models (LLMs) opens new frontiers for unmanned aerial vehicle (UAVs), yet existing  systems remain confined to predefined tasks due to hardware-software co-design challenges. This paper presents the first aerial intelligent agent capable of open-world task execution through tight integration of LLM-based reasoning and robotic autonomy. Our hardware-software co-designed system addresses two fundamental limitations: (1) Onboard LLM operation via an edge-optimized computing platform, achieving 5-6 tokens/sec inference for 14B-parameter models at 220W peak power; (2) A bidirectional cognitive architecture that synergizes slow deliberative planning (LLM task planning) with fast reactive control (state estimation, mapping, obstacle avoidance, and motion planning). Validated through preliminary results using our prototype, the system demonstrates reliable task planning and scene understanding in communication-constrained environments, such as sugarcane monitoring, power grid inspection, mine tunnel exploration, and biological observation applications. This work establishes a novel framework for embodied aerial artificial  intelligence, bridging the gap between task planning and robotic autonomy in open environments.
	\end{abstract}
	
	\section{Introduction}
	
	Although UAVs have advanced significantly, they are still mostly limited to performing tasks that are specifically designed and pre-programmed. Typical applications of UAVs include precision agriculture, power grid inspection, mining industry, biological observations, etc. With the rise of LLMs and visual-language models (VLMs), an important question arises: Can UAVs carry out tasks based on open-set language instructions and interact with open-set objects? 
	Is it possible in a few-shot even a zero-shot manner?
	We refer to such LLM-empowered UAVs that can perform general-purpose tasks in open-world as \emph{aerial intelligent agents (AIA)}.
	
	To build an aerial intelligent agent, the integration of fast reactive behaviors and slow deliberative reasoning remains a critical challenge, particularly in open environments requiring real-time responsiveness and a long-horizon planning. Inspired by the dual-process theory of human cognition~\cite{kahneman2011thinking}, this paper proposes a hardware-software co-designed framework. It embeds fast thinking for agile reaction and slow reasoning for long-horizon task-planning into aerial intelligent agents.
	
	There are several challenges to implement an aerial intelligent agent. First, LLMs are characterized
	by their vast parameter sizes and training on internet-scale
	datasets. Due to limitations in weight, power consumption, and size of UAVs, designing an appropriate edge computing platform and selecting an proper LLM is crucial. Current UAVs designed for specific tasks do not have the computing power to run an LLM onboard. Although it is possible to run the LLM on a ground station and communicate with the UAV wirelessly, this approach is limited by bandwidth and introduces risks such as communication instability and latency. Moreover, it is unsuitable for general-purpose tasks in open world that might have poor communication conditions.
	Second, there is a lack of systematic guidelines for prompt design that integrating the slow deliberative reasoning ability of LLMs  and fast reactive abilities. Common reaction includes state estimation~\cite{xu2022fast}, perception, obstacle avoidance, motion planning~\cite{ren2025safety}, and control.
	
	Recent efforts to integrate LLMs with UAVs predominantly focus on cloud-dependent architectures or offline task planning~\cite{javaid2024large}, which fundamentally constrain real-time adaptability in open environments. While edge computing solutions have emerged for ground robots~\cite{kim2024survey}, their direct adaptation to aerial platforms remains impractical due to stringent power-weight tradeoffs and vibration-induced computational instability. 	
	In this paper, to enable the implementation of aerial intelligent agents, we present a hardware platform and a prompt-design framework for LLM-empowered UAVs. 
	The major contributions in this work are summarized as follows:
	\begin{itemize}
		\item A novel hardware-software co-design framework featuring an edge computing for an UAV prototype. The embedded computing module enables efficient LLM inference (14B parameters DeepSeek-R1 LLM \cite{guo2025deepseek} or 11B parameters Meta Llama 3.2 Vision \cite{dubey2024llama} at 5-6 tokens/second) while maintaining power efficiency ($220$W peak consumption).
		\item We propose a prompt design framework that integrates LLM-based reasoning with conventional UAV autonomy modules. Our architecture achieves bidirectional integration between high-level task planning (via LLMs) and low-level reaction pipelines (state estimation, motion planning, control, etc).
		\item Using the prototype of our hardware platform, we validate the effectiveness of LLMs/VLMs running onboard. We test the performance through mission-critical applications, including sugarcane monitoring, power grid inspection, mine tunnel exploration, and biological observation, demonstrating the generalization ability of task planning and scene understanding in real-world scenarios.
	\end{itemize}
	
	\begin{figure*}[t]
		\centering
		\includegraphics[width=0.9\textwidth]{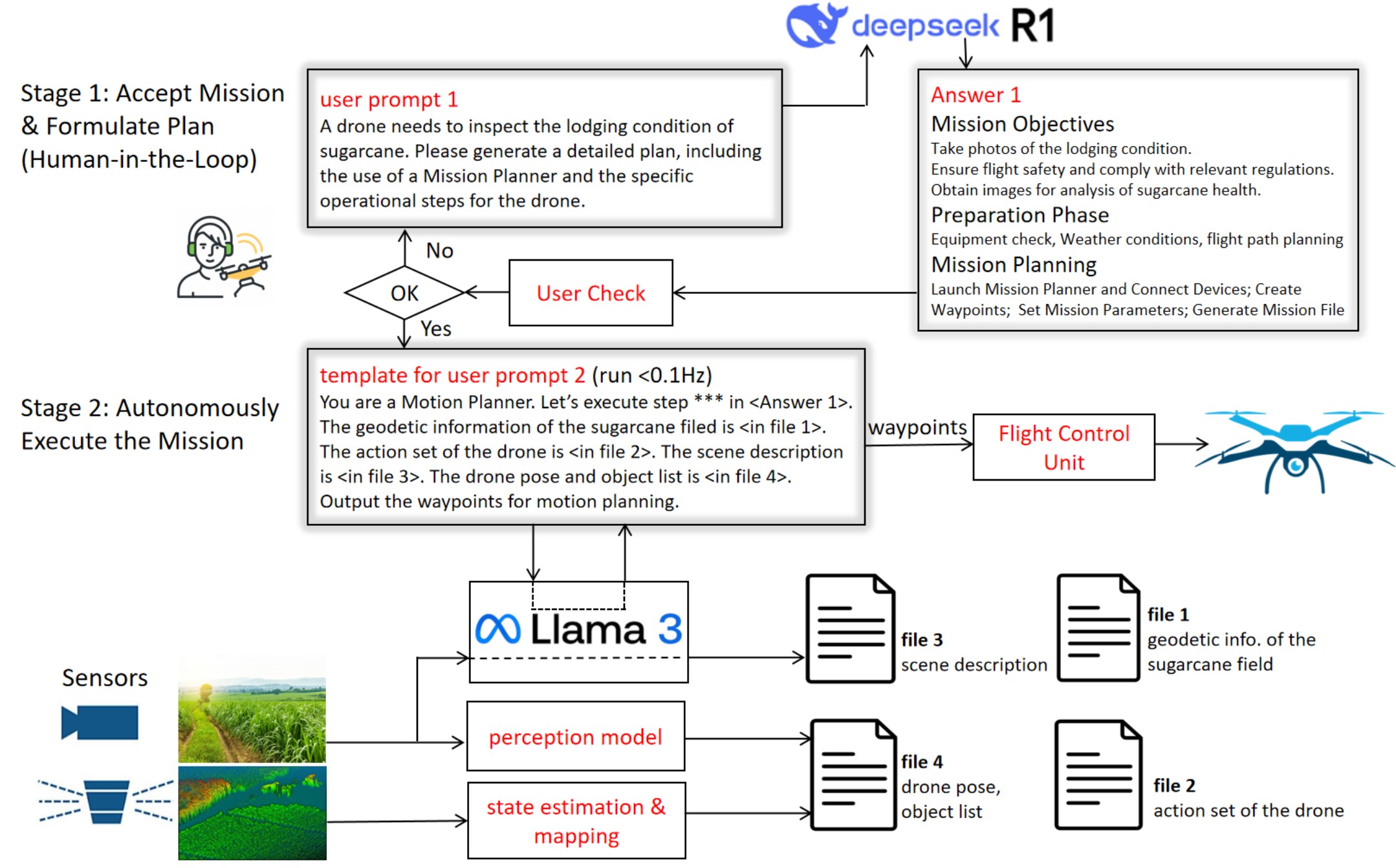}
		\caption{Work flow and prompt design of the proposed aerial intelligent agent.} 
		\label{fig:flow}
	\end{figure*}
	
	\section{Related Work}
	In recent years, the integration of LLMs has revolutionized the field of robotics, enabling robots to communicate, understand, planning, and reason with human-like instruction.
	There are many works that integrate LLM with robots and UAVs \cite{javaid2024large,kim2024survey}. We present some representative works.
	
	LLMs have significantly advanced the field of natural language processing (NLP) since the ChatGPT series~\cite{radford2018improving}, enabling a wide range of applications from text generation to code generation and mathematical reasoning. VLMs~\cite{dubey2024llama} have emerged as a powerful tool for integrating visual and language processing, enabling tasks such as visual question answering and scene understanding. VLMs have shown significant improvements in multimodal tasks, providing more comprehensive and context-aware responses. 
	These models have potential to a wide range of applications, including autonomous vehicles and intelligent customer service systems, by facilitating more natural and efficient interactions between humans and machines.
	Recently, DeepSeek-R1~\cite{guo2025deepseek} is an LLM known for its advanced capabilities in text generation, code generation, and mathematical reasoning. This model has achieved state-of-the-art or competitive performance on various benchmarks.
	
	In \cite{fan2022aerial}, a task of aerial vision-and-dialog navigation was proposed to navigate a drone via natural language conversation. 
	In \cite{liu2023aerialvln}, a task named AerialVLN was proposed, which is UAV-based and towards outdoor environments. The visual language navigation (VLN) is the process of guiding an agent to navigate through an environment based on natural language instructions and visual inputs.
	In \cite{zhao2025cityeqa}, a novel agent tailored for embodied question answering was proposed, which enables long-horizon planning and hierarchical task execution. Recently, some datasets for UAV's VLN have been proposed in \cite{lee2024citynav,gao2024embodiedcity}. Swarm-GPT enables users to automatically generate synchronized drone performances through natural language instructions \cite{jiao2023swarm}. The TPML method~\cite{cui2024tpml} uses LLMs as a command interface to comprehend operators' intentions and translate them into executable codes.  However, LLMs run on cloud services or ground stations in these works.
	
\begin{figure*}[h!]
	\centering
	\begin{minipage}[b]{0.65\textwidth}
		\centering
		\includegraphics[width=\textwidth]{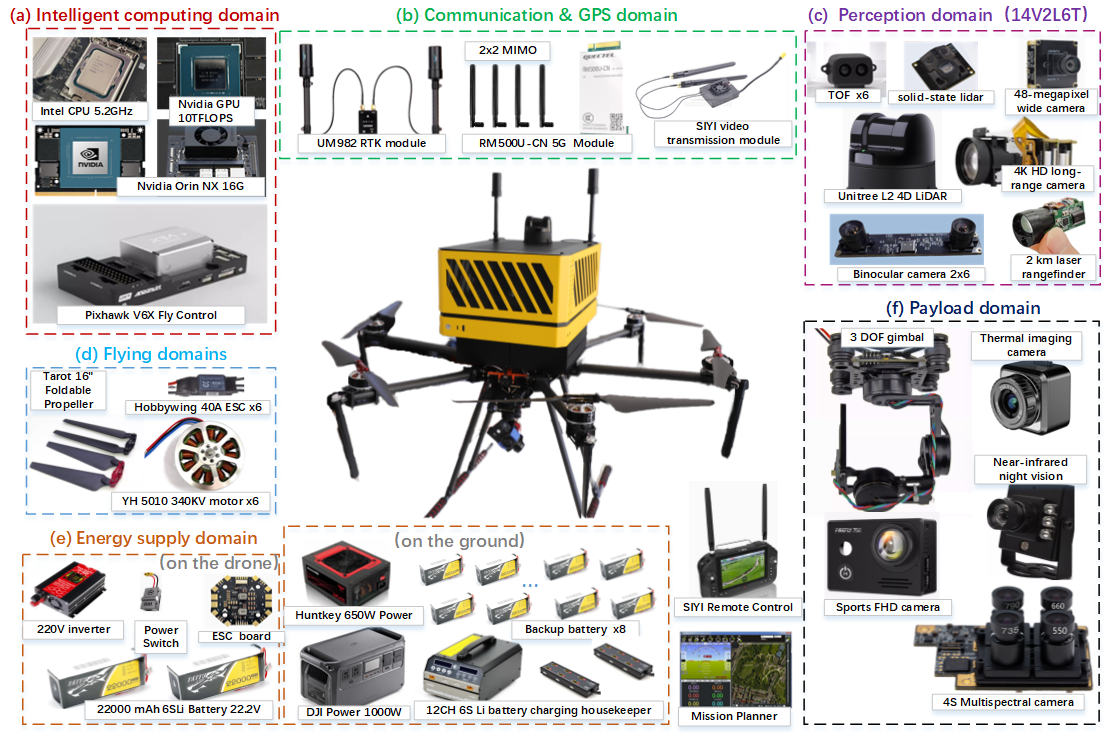}
		\subcaption{Hardware prototype of the aerial intelligent agent and its core components. 
		}
	\end{minipage}
	\hfill
	\begin{minipage}[b]{0.3\textwidth}
		\centering
		\begin{minipage}[b]{\textwidth}
			\centering
			\includegraphics[width=\textwidth]{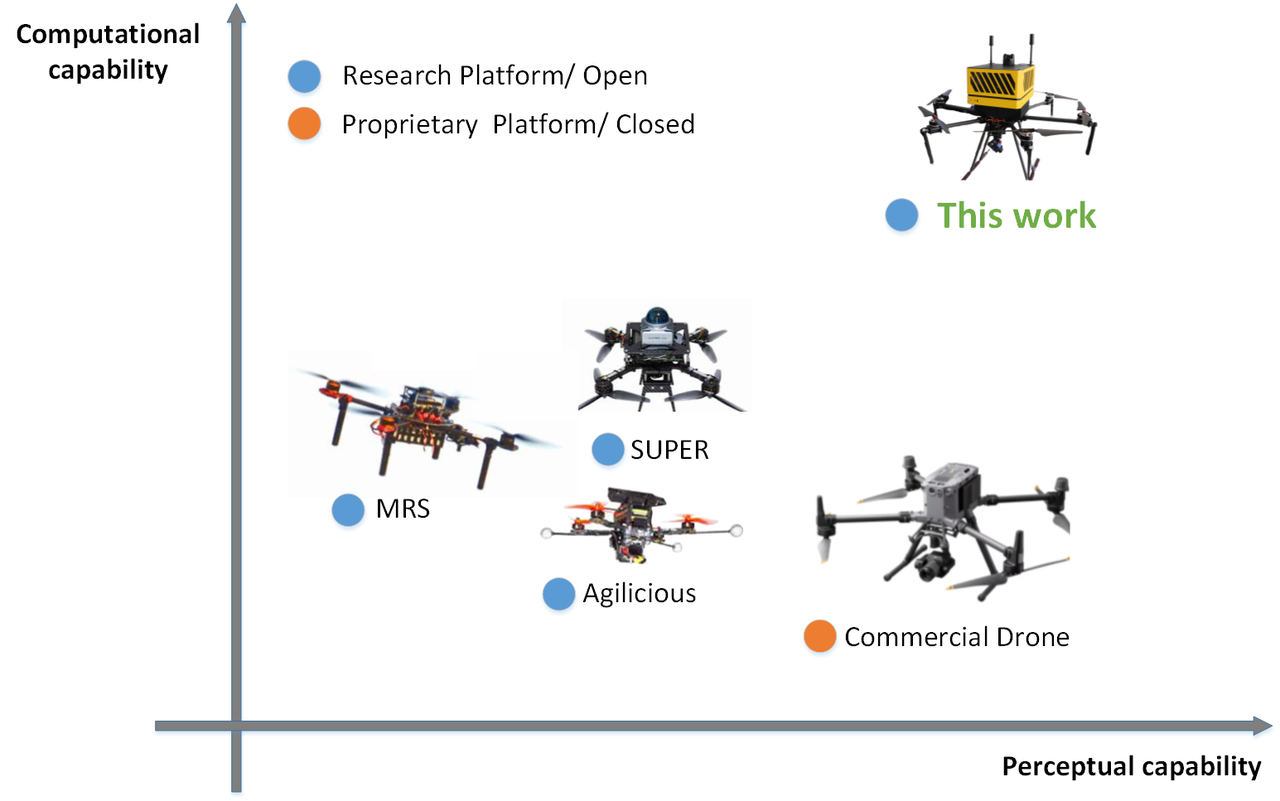}
			\subcaption{perceptual capability and computational capability}
		\end{minipage}\\
		\begin{minipage}[b]{\textwidth}
			\centering
			\includegraphics[width=\textwidth]{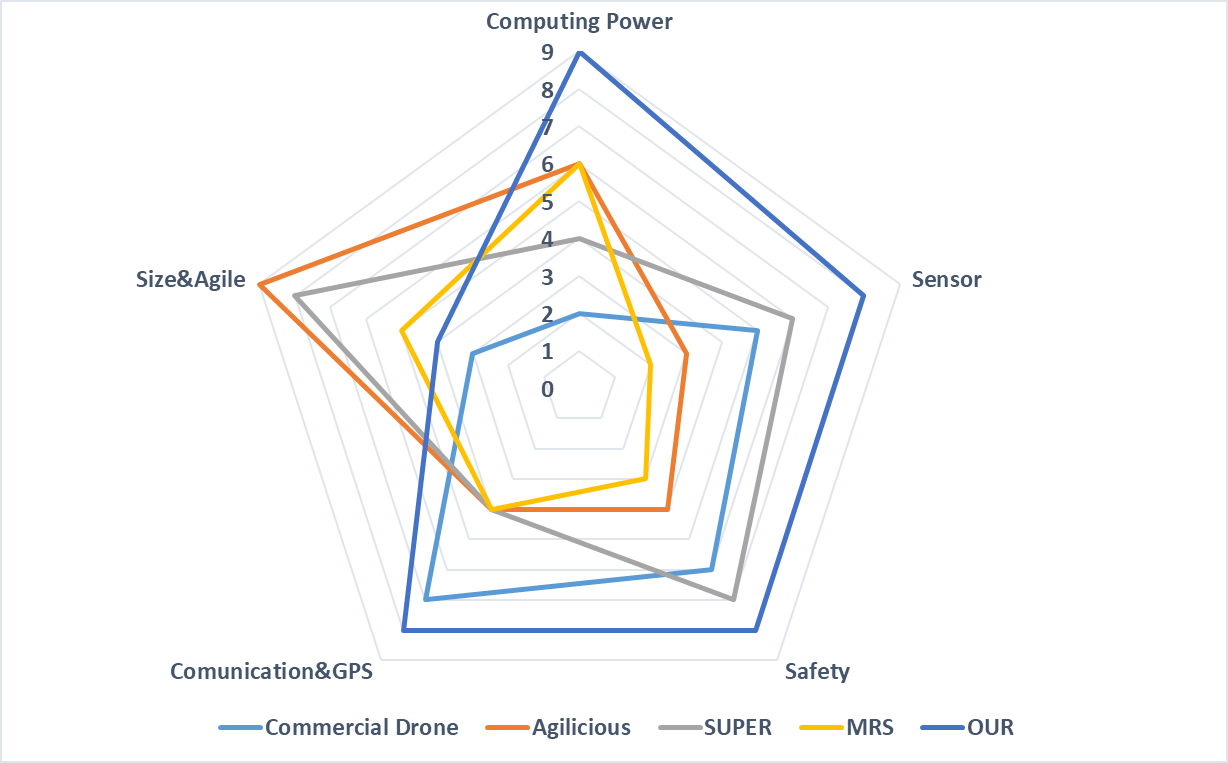}
			\subcaption{radar chart}
		\end{minipage}
	\end{minipage}
	
	\caption{Hardware prototype of the aerial intelligent agent, and the comparison with other state-of-the-art UAVs, including a representative commercial drone DJI 350RTK~\cite{DJI}, MRS~\cite{baca2021mrs}, the open-source agile quadrotor platform (Agilicious) \cite{foehn2022agilicious}, and SUPER~\cite{ren2025safety}.}
	\label{fig:prototype}
\end{figure*}
	
	\section{Framework}

	The core ability of aerial intelligent agent is long-horizon planning for general-purpose tasks in open environments. First of all, we need to clarify two levels of planning, i.e., task planning and motion planning.
	The whole planning procedure involves organizing actions to solve given tasks,
	typically through generating a sequence of high-level
	symbolic operators (i.e., \emph{task planning}) followed by executing them using low-level \emph{motion planning}.
	Motion planning refers to the process of generating a path
	by computing sequential waypoints within configuration or
	task spaces.
	
	The work flow of the aerial intelligent agent is shown in Fig.~\ref{fig:flow}. The prompt design for LLMs/VLMs is the key to unleash the power of LLMs.
	Our method includes two stages.	
	The first stage involves accepting the mission and formulating the task plan, with human-in-the-loop operations. At this stage, we pose a question to the LLM (DeepSeek-R1~\cite{guo2025deepseek} is used in this paper) via a prompt, providing it with a description of the task and requesting it to generate a detailed plan. The answer provided by the LLM is then reviewed and confirmed by a human operator. If the plan is deemed reasonable, the process proceeds to the next stage. If not, the prompt is rerun to generate a revised response. Drawing on the vast knowledge acquired through LLMs, it possesses common sense and can provide rational planning for the mission, even if some details are not mentioned in the prompt. The common sense is considered one of the keys to achieving machine intelligence \cite{kejriwal2024nature}.
	
	The second stage involves the UAV autonomously executing the mission. We take the sugarcane monitoring as an example. The images, LiDAR point clouds, and GNSS/IMU data collected by the UAV are sent to the perception, state estimation and mapping modules.
	The perception module can be a small deep-learning model to detect obstacles, such as high-voltage power lines, buildings, etc.
	The state estimation and mapping module is used to provide the ego pose of the UAV and a local point cloud map. This module can be implemented by a LiDAR-inertial odometry. We use Fast-LIO2~\cite{xu2022fast} in our system.
	Then the UAV's ego pose, local map, and the obstacle detection results are fused and saved into a file. It provides the UAV's pose and the attributes of the obstacle, including category, position, distance, etc. 
	
	At the second stage, the images are also sent to a VLM to generate scene descriptions of the images, which are also written into a file. In addition, there are two fixed files that record the geodetic information of the sugarcane field and the action set of the UAV, respectively. The content from these four files is substituted into a text template, forms a prompt that allows the LLM to act as a Motion Planner. It makes decision by selecting an action from the action set and generates waypoints for motion planning. The generated waypoints are input into the flight control unit (FCU) to produce commands for controlling the UAV. The VLM does not require high-frequency execution; it is typically performed every few seconds to dozens of seconds. In this paper, the Meta Llama 3.2 Vision 11B~\cite{dubey2024llama} is used for both VLM and LLM at this stage.
	
	The action set of the UAV contains three main parts. (1) Basic flight control, including takeoff, land, hover, emergency stop. (2) Three-axis motion control, including linear motion and angular motion. (3) Dynamic parameter control, including accelerate, decelerate, altitude hold, speed lock.
	In addition there are some advanced modes, including waypoint navigation, return to home (RTH), orbit, etc.
		
	\section{Hardware Design}

	The hardware platform of the aerial intelligent agent is shown in Fig.~\ref{fig:prototype}(a). 
	Its mechanical structure forms a hexacopter drone, which is called \emph{Bumblebee LLM UAV} 1.0.
	It has a 850-millimeter wheelbase. The weight is between 10 and 15 kg, depending on the payloads.
	Drawing inspiration from the human body structure and its functions, the hardware can be divided into 6 domains as introduced below.

	\textbf{Intelligent Computing Domain (Brain)}: 
	Our UAV is equipped with a combination of a 16-core 5GHz Intel CPU and a 12-core 2.5GHz ARM processor. It is also equipped with a dual-layer GPU architecture, capable of delivering up to 10 TFLOPS of general-purpose computing power and 40 TOPS (Int8) of embedded computing power. Our UAV features a substantial 96GB DDR5 memory. Additionally, its NVIDIA Jetson Orin module is equipped with 16GB of  LPDDR memory and 8GB of high-speed GDDR memory, further enhancing its overall performance capabilities.
	The hardware and software platform is capable of running the DeepSeek 14B-parameter model at a speed of 5-6 tokens per second (CPU mode), with a power consumption of approximately 220W.
	
	\textbf{Energy Supply Domain (Heart)}: This domain contains the battery compartment and power supply system. Currently, it can provide a peak power consumption of 1000Wh to the intelligent computing domain, allowing continuous operation of large models for 4 hours. 
	It can also provide 22000WmAh (or 700Wh) for the flight domain.
	A mobile power supply platform (mobile charging station) is provided, capable of charging up to 12 batteries simultaneously, meeting the daily work requirements.
	
	\textbf{Flying Domain (Body, Skeleton, and Limbs)}: This domain includes a FCU and propellers. Each propeller can provide a thrust of 3kg, and with a 6-propeller configuration, it can lift a maximum weight of 18kg.
	
	\textbf{Communication \& GPS Domain (Ears and Mouth)}: The 2$\times$2 MIMO 5G communication module is equipped and can provide a theoretical bandwidth of 2Gbps, with an actual bandwidth of 400Mbps. This is 10 times the communication bandwidth of current commercial drone data transmission modules. The RTK module can provide centimeter-level navigation accuracy. The video transmission module can achieve a transmission distance of 30 km, which is a 30\% increase compared to the DJI M350 RTK (with a transmission distance of 20 km) ~\cite{DJI}.
	
	\textbf{Perception Domain (Eyes and Tactile Sense)}: This domain is equipped with the following sensors, which can be abbreviated as 14V+6T+1L.
	\begin{itemize}
		\item 14 cameras (14V). It includes 6 stereo camera systems, one wide-angle camera with 48 million pixels, and one 4K ultra-high-definition telephoto camera;
		\item 6 Time-of-Flight (ToF) cameras (6T);
		\item 4D LiDAR (1L);
		\item An IMU. The FCU integrates with an IMU.
	\end{itemize}

	\textbf{Payload Domain (Work Tools)}: It is equipped with a diverse range of payloads, including one infrared thermal camera, laser rangefinder (1.5KM), 1080p FPV camera, 3-axis gimbal, searchlight (for night operations), wide-angle night vision device, telephoto night vision device, wide-angle camera, telephoto camera, etc.
	
	\begin{figure}[t]
		\centering
		\includegraphics[width=0.48\textwidth]{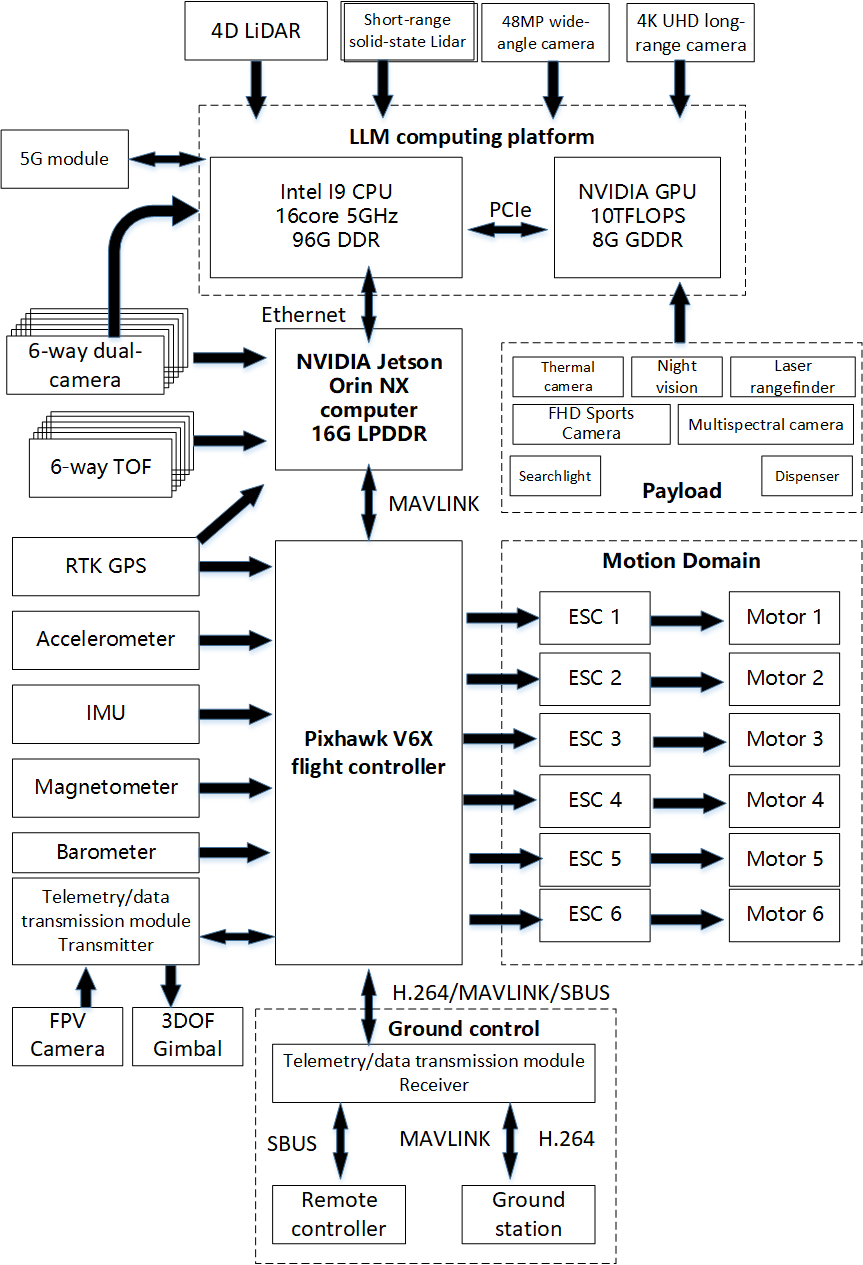}
		\caption{Flowchart of the hardware platform.
		} 
		\label{fig:framwork}
	\end{figure}
	
	The flowchart of the hardware platform is shown in Fig.~\ref{fig:framwork}.
	All hardware components in our prototype are off-the-shelf products. As a result, the hardware platform can ensure performance and reliability. Additionally, it demonstrates significant cost advantages by eliminating the need to pay the high Non-Recurring Engineering (NRE) fees associated with customized components. 
	
	The comparison of our prototype with state-of-the-art UAVs is shown in Fig.~\ref{fig:prototype}(b)(c). It can be seen that different UAV products have their own emphases in design philosophy and application scenarios. The SUPER~\cite{ren2025safety} has advantages in computing power and safety. The Agilicious \cite{foehn2022agilicious} has its own strength in power consumption control and embedded applications. DJI drones~\cite{DJI} stand out in sensor configuration and positioning capabilities, making them particularly suitable for aerial photography. By contrast, our prototype has comprehensive advantages in computing power, sensing, safety, and communication capabilities, and can meet the needs of a variety of complex mission scenarios.
	
	Our prototype and DJI drones are weak in size and agility, primarily due to the following reasons. First, to achieve high endurance, large-capacity batteries are required to provide sufficient propulsion, and larger lift is needed to overcome their own weight, which in turn necessitates bigger propellers. Second, in open fields and over the sea, considering the impact of wind, UAVs need to have enough power and weight to counteract it and ensure stable flight. Lastly, powerful computing platforms require extensive cooling measures. A compact design may lead to ineffective heat dissipation, thereby affecting operational stability.
	
	The key innovations of the our hardware platform is summarized below. First, for the first time in 10-15Kg class UAVs, an hardware platform for running LLMs has been introduced, supporting a 14B-parameter LLM at the edge. The computing power is enhanced by 2-3 orders of magnitude compared to existing commercial UAV platforms.
	Second, the platform is equipped with cameras, ToF LiDARs, and 4D LiDAR modules to enable real-time perception, improving the UAV's obstacle avoidance capabilities and its potential to operate autonomously in challenging weather conditions.
	The hardware of our prototype will be open. 
	
	\section{Experiments}
	
	\begin{figure}[t]
		\centering
		\includegraphics[width=0.49\textwidth]{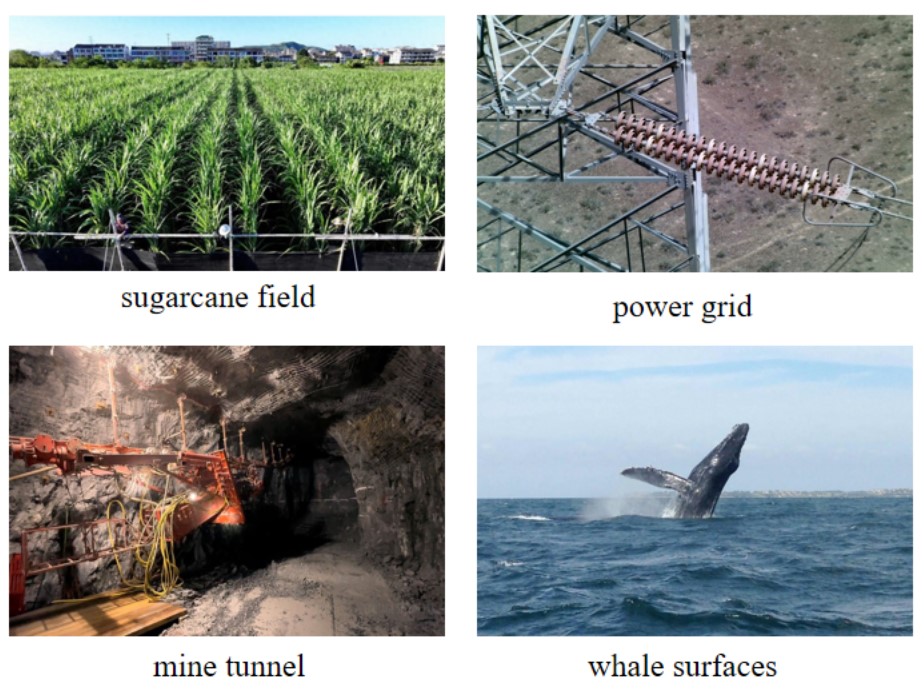}
		\caption{Applications of the aerial intelligent agent, including sugarcane monitoring, power grid inspection, mine tunnel exploration, and biological observation of whale surfaces.
		} 
		\label{fig:app}
	\end{figure}
	
	We implement a prototype of an aerial intelligent agent based on the proposed prompt design method and hardware platform.
	We evaluate its task planning and scene understanding abilities using its onboard computing.
	To demonstrate its effectiveness to deal with complex tasks, the performance is evaluated for four different tasks, including sugarcane monitoring, power grid inspection, mine tunnel exploration, and biological observation of whale surfaces, as shown in Fig.~\ref{fig:app}. 
	
	\subsection{Task Definition}
	
	\emph{Sugarcane Monitoring}: In the field of agricultural breeding, aerial intelligent agents have the potential to utilize their high-precision measurement and perception capabilities to conduct lodging measurement and growth monitoring of sugarcane. Through the collaborative operation of fast-thinking and slow-thinking systems, agents are able to autonomously identify the growth status of sugarcane, promptly detect lodging and other issues, and provide strong support for agricultural industrialization.
	
	\emph{Power Grid Inspection}: Aerial intelligent agents have the potential to achieve autonomous flight following and defect recognition. The fast-thinking system is responsible for obstacle avoidance and path planning, ensuring that the aerial agent flies safely and efficiently along the power lines. Meanwhile, the slow-thinking system conducts in-depth analysis of the inspection data, identifies potential defects, and provides crucial references for the maintenance of power facilities. 
	
	\emph{Mine Tunnel Exploration}: Aerial intelligent agents can leverage their LiDAR SLAM technology for precise navigation and mapping. By integrating the real-time perception and path-planning capabilities of the fast-thinking system \cite{ren2025safety} with the comprehensive analytical and processing power of the slow-thinking system, UAVs are able to autonomously traverse complex mine environments and conduct all-round monitoring. This provides a solid basis for the safe exploitation and sustainable utilization of mine tunnels.
	
	\emph{Biological Observation of Whale Surfaces}: Rendezvous with sperm whales for biological observations is made challenging by their prolonged dive patterns \cite{jadhav2024reinforcement}. Whale sightings at the surface, known as rendezvous, are critical for connecting behavioral data with tagged whales but can easily be missed.
	The fast-thinking system employs real-time multi-sensor fusion to maintain good positional accuracy during whale surfacing events. The slow-thinking system conducts behavioral forecasting, integrating historical dive patterns with oceanographic data to predict the surfacing zones.
	
	\subsection{Task Planning Results at the First Stage}
	
	Task planning results at the first stage for different applications are shown in the following four boxes. We can see that the LLM~\cite{guo2025deepseek} can generate reasonable task planning for all the applications, including mission objectives, preparation phase, mission planning, etc. 
	Meanwhile, certain steps in task planning vary according to the specific task at hand and incorporate common sense. Take the takeoff point selection as an example, sugarcane monitoring application selects an open area near the sugarcane field as the takeoff point. By contrast, tunnel exploration application sets the takeoff point at the entrance of the mine tunnel and ensure that there are no obstacles in the area. In addition, the LLM noticed that there is no GPS signal inside the mine tunnels, even though we did not mention this point in the prompt. Please refer to the red text in the box below. We can see that the reasoning abilities of LLMs enable task planning to address a wide range of general-purpose tasks effectively.
	
\begin{mybox}[Application 1: Sugarcane Monitoring]
	\scriptsize
	\textbf{Prompt}: A drone needs to inspect the lodging (fallen over) condition of sugarcane. Please generate a detailed plan, including the use of a Mission Planner and the specific operational steps for the drone.
	
	\textbf{Answer}: 
	
	Mission Objectives
	\begin{itemize}
		\item Take photos or videos of the lodging condition in the sugarcane field.
		\item Ensure flight safety and comply with relevant regulations.
		\item Obtain high-resolution images for subsequent analysis of sugarcane health.
	\end{itemize}
	
	Preparation Phase
	\begin{itemize}
		\item Equipment check, including battery level, remote controller, mission planner software, GPS signal.
		\item Weather conditions. Ensure that the weather is clear, with wind speeds not exceeding Beaufort Scale 4. Avoid flying on rainy days or in environments with high humidity to prevent any adverse effects on the equipment's performance.
		\item Site Planning: (1) Determine the location and size of the sugarcane field; (2) Plan the drone’s flight path to avoid obstacles (such as power lines, buildings, etc.).
	\end{itemize}
	
	Mission Planning (Using Mission Planner)
	\begin{itemize}
		\item Launch Mission Planner and Connect Devices. Launch the Mission Planner software.
		Connect the drone and the remote controller to the computer, ensuring that the telemetry data is displayed properly.
		\item Create Waypoints. (1) Select an open area near the sugarcane field as the takeoff point. (2) Plan the Flight Path, including Add waypoints, Set the flight altitude, and Adjust the flight speed. (3) Mark the Landing Point.
		\item Set Mission Parameters, including photo interval, video recording, and automatic return-to-home (RTL or RTH) point.
		\item Generate Mission File.
	\end{itemize}
\end{mybox}

\begin{mybox}[Application 2: Power Grid Inspection]
	\scriptsize
	\textbf{Prompt}: A drone needs to conduct inspections on the power grid, and the inspection area is in the mountainous region. Please provide a detailed plan, including the use of Mission Planner and the specific operating procedures for the drone.
	
	\textbf{Answer}: 
	
	Mission Objectives
	\begin{itemize}
		\item Inspect the condition of transmission lines, towers, and other power facilities.
		\item Identify potential issues (such as damage, loose components, etc.) and record them.
		\item Ensure flight safety and comply with relevant regulations.
	\end{itemize}
	
	Preparation Phase
	\begin{itemize}
		\item Equipment check, including battery level, remote controller, mission planner software, GPS signal.
		\item On-site Reconnaissance. (1) Create a detailed map of the flight area, marking the locations of transmission lines, towers, and obstacles. (2) Identify no-fly zones. (3)
		Consider weather conditions to ensure flight safety.
	\end{itemize}
	
	Mission Planning (Using Mission Planner)
	\begin{itemize}
		\item Flight Path Design. (1) Import a topographic map using Mission Planner software and mark the locations of the power transmission lines and towers. (2) Design a low-altitude inspection flight path to ensure coverage of all key areas.
		(3) Set obstacle avoidance points and detour routes for complex terrain areas.
		\item Waypoint Settings. (1) Add takeoff and landing points, inspection points, and emergency landing points in Mission Planner. (2) Set waypoints for the drone to hover and take photos or record videos near the power transmission lines.
		\item Flight Parameter Adjustment. (1) Set an appropriate flight speed. (2) Adjust the camera angles. (3) Set the Return-to-Home (RTH) point for low battery conditions.
		\item Signal Backup. Set up relay devices or backup communication links in Mission Planner to prevent signal loss in mountainous areas.
		\item Battery Management.
	\end{itemize}
\end{mybox}

	\begin{figure*}[t]
		\centering
		\includegraphics[width=0.9\textwidth]{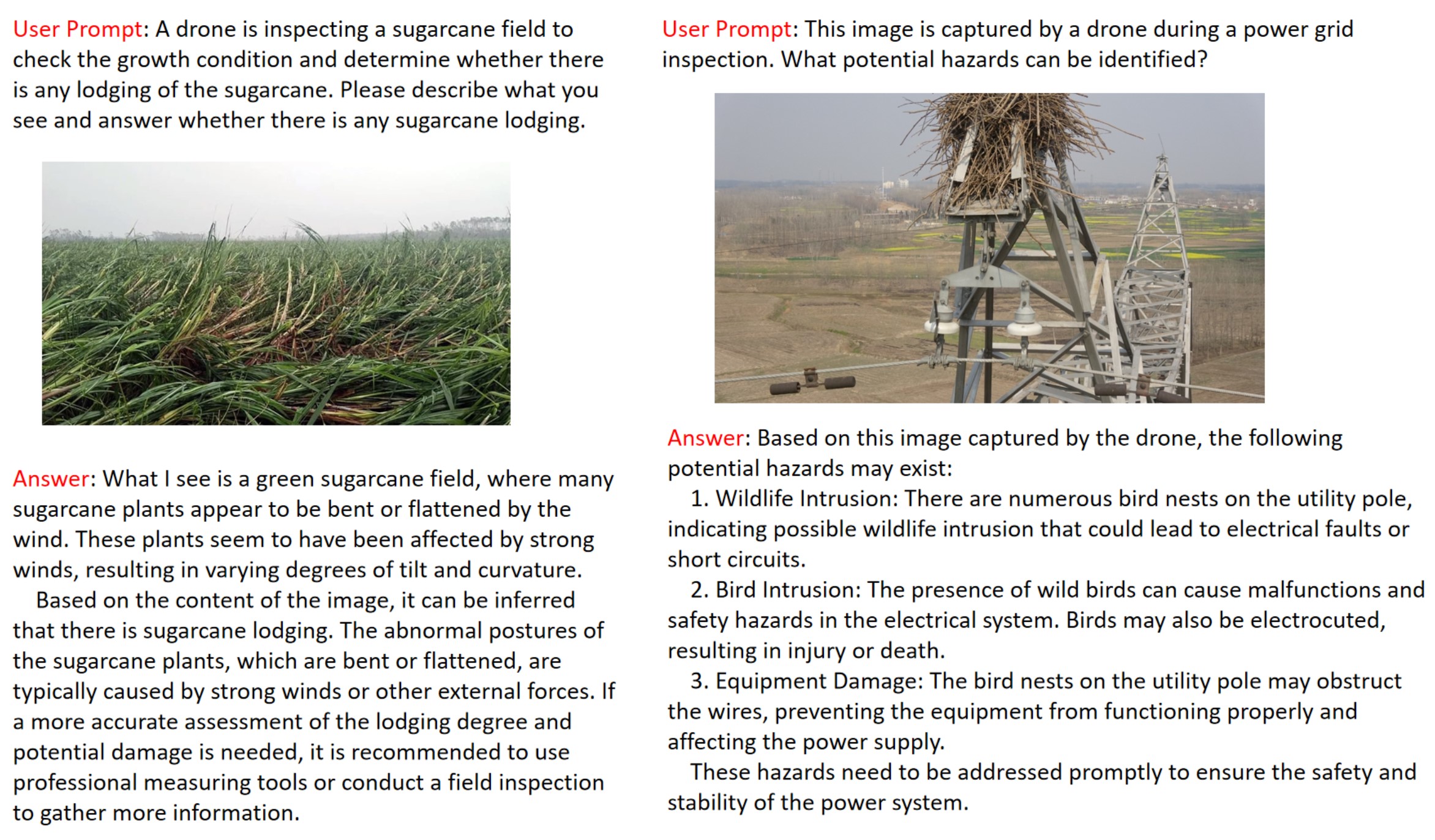}
		\caption{Scene description of sugarcane monitoring and power grid inspection using a VLM.
		} 
		\label{fig:corner}
	\end{figure*}

\begin{mybox}[Application 3: Mine Tunnel Exploration]
	\scriptsize
	\textbf{Prompt}: A drone needs to model the conditions inside a mine tunnel using LiDAR. There is no light inside the tunnel. Please provide a detailed plan, including the use of Mission Planner and the specific operating procedures for the drone.
	
	\textbf{Answer}: 
	
	Mission Objectives
	\begin{itemize}
		\item Utilize a drone equipped with LiDAR to complete the acquisition of 3D point cloud data inside the mine tunnel and generate a high-precision model of the tunnel.
	\end{itemize}
	
	Preparation Phase
	\begin{itemize}
		\item Hardware Preparation. (1) Equip the drone with a LiDAR. (2)	Include a high-precision GPS module \textcolor{red}{(RTK positioning enhancement can be used if the GPS signal is weak inside the mine tunnel)}. (3) Use a drone that supports flight in no-light environments (such as DJI Phantom Pro or other professional models).
		\item Software Preparation. (1) Install and configure the Mission Planner software. \textcolor{red}{Set ``GPS\_TYPE'' to ``None'' because there is no GPS signal in the mine tunnel.} (2) Ensure that the communication between the LiDAR device and the drone is functioning properly.
	\end{itemize}
	
	Mission Area Analysis
	\begin{itemize}
		\item Tunnel Dimensions: Clarify the specific length, width, and height of the mine tunnel (it is recommended to conduct a preliminary survey). 
		\item Terrain Complexity: Identify if there are branches, corners, or obstacles.
		\item Safety Exit: Ensure there are emergency landing points along the flight path.
	\end{itemize}
	
	Mission Planning (Using Mission Planner)
	\begin{itemize}
		\item Flight Path Planning, including settings for (1) flight path, (2) altitude, (3) waypoint spacing and overlap, and (4) takeoff and landing points.
		\item Flight Parameter Settings, including settings for (1) flight speed, (2) photo interval, (3) return-to-home altitude.
	\end{itemize}
	
\end{mybox}

\begin{mybox}[Application 4: Biological Observations]
	\scriptsize
	\textbf{Prompt}: A drone needs to capture the moment when whales surface in the ocean, where wind and rain are present. Please provide a detailed plan, including the use of Mission Planner and specific operating procedures for the drone.
	
	\textbf{Answer}: 
	
	Mission Objectives
	\begin{itemize}
		\item Capture the moment when whales surface.
		\item Obtain high-definition photos and videos for scientific research or documentation.
		\item Ensure the stability and safety of the drone under extreme weather conditions.
	\end{itemize}
	
	Preparation Phase
	\begin{itemize}
		\item Equipment check, including drone, camera, batteries, communication equipment, emergency equipment.
		\item Mission area analysis. (1) Identify the areas where whales are active. (2) Plan the flight route to stay as close as possible to the locations where whales are likely to surface, while maintaining an appropriate distance. (3) Mark emergency landing points: Select locations near the boat or other safe areas.
	\end{itemize}
	
	Mission Planning (Using Mission Planner)
	\begin{itemize}
		\item Flight path planning. (1) Set the takeoff and landing points. (2) Plan a circular or diamond-shaped flight path around the whale activity area. (3) In Mission Planner, set the waypoint interval to ensure the drone can adjust its position in real-time.
		\item Mission parameter settings, including mode selection, return-to-home strategy, and safety distance.
		\item Ground station preparation. (1) Ensure the ground control station is connected stably and monitor the drone's status in real-time. (2) Prepare storage devices for saving the captured photos and videos.
	\end{itemize}
\end{mybox}

	\subsection{Scene Understanding Results at the Second Stage}
	Using real-world images of corner cases, We test the onboard VLM's~\cite{dubey2024llama} scene understanding ability.
	Examples for sugarcane monitoring and power grid inspection are shown in Fig.~\ref{fig:corner}, respectively. We can see that the VLM successfully detect the the lodging condition of a sugarcane field and potential hazards of the power grid.
	
	From these experiments, the aerial intelligent has shown its potential to replace humans in performing tasks that are dangerous or monotonous. Our results has two implications. First, compared to the expensive costs of human operators, aerial intelligent agents have enabled the realization of an agent with common sense using relatively low-cost hardware. Second, considering VLMs and LLMs typically do not require high-frequency execution, a UAV can be powered with one day of slow thinking capability using just one kilowatt-hour of electricity. Our solution thus offers great potential, providing plenty of room for a wide range of applications

	\section{Conclusion}
	We present a framework for aerial intelligent agents through the integration of LLMs with UAV systems. The proposed framework achieves three critical advancements: 1) Edge deployment of 14B-parameter LLMs via optimized hardware design, overcoming traditional power-compute constraints; 2) A two-stage prompt design framework enabling effective interaction between LLM-guided task planning and low-level reaction abilities; 3) Effectiveness in mission-critical applications requiring both environmental awareness and operational precision.
	Our experiments reveal two key insights: First, the bidirectional information flow between LLM reasoning and reaction modules can be applied to general-purpose tasks with specific-design methods. Second, onboard LLM running is beneficial for UAVs, especially for safety-critical operations.

	In the future, we will further reduce the UAV weight. The current computing platform uses an aluminum alloy frame. In the future, it can be replaced with lighter materials such as carbon fiber, which have the same level of rigidity. We will further validate the effectiveness of the system through real flight tests, and will make our hardware platform open.

	{\small
		\bibliographystyle{IEEEbib}
		\bibliography{egbib}
	}

\end{document}